\documentclass[runningheads]{llncs}
\usepackage[T1]{fontenc}
\usepackage{caption}
\usepackage{cite}
\usepackage[symbol]{footmisc}
\usepackage{graphicx}
\usepackage{amsmath}
\usepackage{amssymb}
\usepackage{makecell}
\usepackage{float}
\usepackage{multirow}
\usepackage{booktabs}
\newcommand{\sslash}{\mathbin{/\mkern-4mu/}}
\usepackage[colorlinks,linkcolor=blue]{hyperref}
\usepackage{color}

\begin{document}
\title{Efficient Subclass Segmentation in Medical Images}
\author{Linrui Dai\inst{1}\and Wenhui Lei\inst{1,2}\and Xiaofan Zhang\inst{1,2,*}}
\institute{~\inst{1}Shanghai Jiao Tong University\\
~\inst{2}Shanghai Artificial Intelligence Laboratory
\email{\\\{o.o111, wenhui.lei, xiaofan.zhang\}@sjtu.edu.cn}}

\maketitle

\begin{abstract}
 As research interests\footnotetext[1]{Corresponding author.} in medical image analysis become increasingly fine-grained,  the cost for extensive annotation also rises. One feasible way to reduce the cost is to annotate with coarse-grained superclass labels while using limited fine-grained annotations as a complement. In this way, fine-grained data learning is assisted by ample coarse annotations. Recent studies in classification tasks have adopted this method to achieve satisfactory results. However, there is a lack of research on efficient learning of fine-grained subclasses in semantic segmentation tasks. In this paper, we propose a novel approach that leverages the hierarchical structure of categories to design network architecture. Meanwhile, a task-driven data generation method is presented to make it easier for the network to recognize different subclass categories. Specifically, we introduce a Prior Concatenation module that enhances confidence in subclass segmentation by concatenating predicted logits from the superclass classifier, a Separate Normalization module that stretches the intra-class distance within the same superclass to facilitate subclass segmentation, and a HierarchicalMix model that generates high-quality pseudo labels for unlabeled samples by fusing only similar superclass regions from labeled and unlabeled images. Our experiments on the BraTS2021 and ACDC datasets demonstrate that our approach achieves comparable accuracy to a model trained with full subclass annotations, with limited subclass annotations and sufficient superclass annotations. Our approach offers a promising solution for efficient fine-grained subclass segmentation in medical images. Our code is publicly available \href{https://github.com/OvO1111/EfficientSubclassLearning}{here}.

\end{abstract}
\section{Introduction}
In recent years, the use of deep learning for automatic medical image segmentation has led to many successful results based on large amounts of annotated training data. However, the trend towards segmenting medical images into finer-grained classes (denoted as $subclasses$) using deep neural networks has resulted in an increased demand for finely annotated training data\cite{bakas2018identifying, sekuboyina2021verse, he2021synergistic}. This process requires a higher level of domain expertise, making it both time-consuming and demanding. As annotating coarse-grained (denoted as $superclasses$) classes is generally easier than subclasses, one way to reduce the annotation cost is to collect a large number of superclasses annotations and then labeling only a small number of samples in subclasses. Moreover, in some cases, a dataset may have already been annotated with superclass labels, but the research focus has shifted towards finer-grained categories\cite{finer1, finer2}. 
In such cases, re-annotating an entire dataset may not be as cost-effective as annotating only a small amount of data with subclass labels. 
%In such cases, it may not be cost-effective to re-annotate the entire dataset, making it more feasible to only annotate a small amount of data for the subclasses.
\par Here, the primary challenge is to effectively leverage superclass annotations to facilitate the learning of fine-grained subclasses. To solve this problem, several works have proposed approaches for recognizing new subclasses with limited subclass annotations while utilizing the abundant superclass annotations in classification tasks \cite{bukchin2021fine, fotakis2021efficient, yang2021towards, ni2021superclass}. In general, they assume the subclasses are not known during the training stage and typically involve pre-training a base model on superclasses to automatically group samples of the same superclass into several clusters while adapting them to finer subclasses during test time.

\par However, to the best of our knowledge, there has been no work specifically exploring learning subclasses with limited subclass and full superclass annotations in semantic segmentation task. Previous label-efficient learning methods, such as semi-supervised learning\cite{yu2019uncertainty, chen2021semi, luo2022semi}, few-shot learning\cite{lei2021one, hansen2022anomaly, ouyang2020self} and weakly supervised learning\cite{kervadec2020bounding, zhang2022cyclemix}, focus on either utilize unlabeled data or enhance the model's generalization ability or use weaker annotations for training. However, they do not take into account the existence of superclasses annotations, making them less competitive in our setting. 

\par In this study, we focus on the problem of efficient subclass segmentation in medical images, whose goal is to segment subclasses under the supervision of limited subclass and sufficient superclass annotations. Unlike previous works such as \cite{bukchin2021fine, fotakis2021efficient, yang2021towards, ni2021superclass}, we assume that the target subclasses and their corresponding limited annotations are available during the training process, which is more in line with practical medical scenarios.

\par Our main approach is to utilize the hierarchical structure of categories to design network architectures and data generation methods that make it easier for the network to distinguish between subclass categories. Specifically, we propose 1) a \textbf{Prior Concatenation} module that concatenates predicted logits from the superclass classifier to the input feature map before subclass segmentation, serving as prior knowledge to enable the network to focus on recognizing subclass categories within the current predicted superclass; 2) a \textbf{Separate Normalization} module that aims to stretch the intra-class distance within the same superclass, facilitating subclass segmentation; 3) a \textbf{HierarchicalMix} module inspired by GuidedMix\cite{Peng2022}, which for the first time suggests fusing similar labeled and unlabeled image pairs to generate high-quality pseudo labels for the unlabeled samples. However, GuidedMix selects image pairs based on their similarity and fuses entire images. In contrast, our approach is more targeted. We mix a certain superclass region from an image with subclass annotation to the corresponding superclass region in an unlabeled image without subclass annotation, avoiding confusion between different superclass regions. This allows the model to focus on distinguishing subclasses within the same superclass. Our experiments on the Brats 2021 \cite{baid2021rsna} and ACDC \cite{bernard2018deep} datasets demonstrate that our model, with sufficient superclass and very limited subclass annotations, achieves comparable accuracy to a model trained with full subclass annotations.

\section{Method}
\subsubsection{Problem Definition}
We start by considering a set of $R$ coarse classes, denoted by $\mathcal{Y}_{c}=\{Y_1,...,Y_R\}$, such as background and brain tumor, and a set of $N$ training images, annotated with $\mathcal{Y}_{c}$, denoted by $\mathcal{D}_{c}=\{(x^l,y^l)|y^l_i\in\mathcal{Y}_{c}\}_{l=1}^N$. Each pixel $i$ in image $x^l$ is assigned a superclass label $y^l_i$. To learn a finer segmentation model, we introduce a set of fine subclass $K=\sum_{i=1}^R{k_i}$ in coarse classes, denoted by $\mathcal{Y}_{f}=\{Y_{1,1},...,Y_{1,k_1},...,Y_{R,1},...,$ $Y_{R,k_R}\}$, such as background, enhancing tumor, tumor core, and whole tumor. We assume that only a small subset of $n$ training images have pixel-wise subclass labels $z\in\mathcal{Y}_f$ denoted by $\mathcal{D}_{f}=\{(x^l,z^l)|z_i^l\in\mathcal{Y}_{f}\}_{l=1}^n$. Our goal is to train a segmentation network $f(x^l)$ that can accurately predict the subclass labels for each pixel in the image $x^l$, even when $n\ll N$. \textbf{Without specification, we consider $R=2$ (background and foreground) and extend the foreground class to multi subclass in this work.}

\subsubsection{Prior Concatenation}
\begin{figure}[tbp!]
\centering
% {left, bottom, right, top}
\includegraphics[page=2, trim={4cm, 5.2cm, 4.2cm, 2.1cm}, clip, scale=0.45]{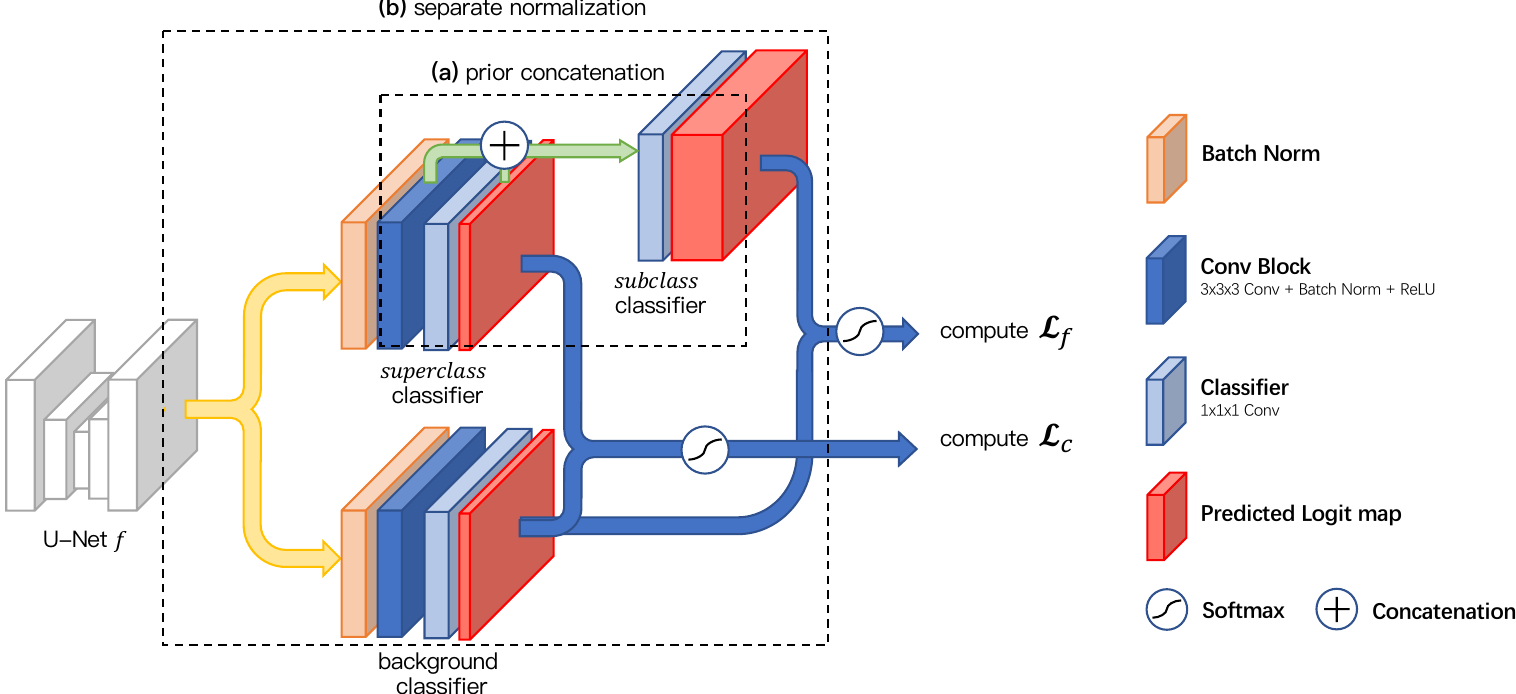}
\caption{Proposed network architecture, $\mathcal{L}_c$ and $\mathcal{L}_f$ stand for the superclass loss and subclass loss respectively.}
\label{fig:Network}
\end{figure}

One direct way to leverage the superclass and subclass annotations simultaneously is using two $1\times1\times1$ convolution layers as superclass and subclass classification heads for the features extracted from the network. The superclassification and subclassification heads are individually trained by superclass $P_c(x^l)$ labels and subclass labels $P_f(x^l)$. With enough superclass labels, the feature maps corresponding to different superclasses should be well separated. 
However, this coerces the subclassification head to discriminate among $K$ subclasses under the mere guidance from few subclass annotations, making it prone to overfitting.

\par Another common method to incorporate the information from superclass annotations into the subclassification head is negative learning \cite{kim2019nlnl}. This technique penalizes the prediction of pixels being in the wrong superclass label, effectively using the superclass labels as a guiding principle for the subclassification head. However, in our experiments, we found that this method may lead to lower overall performance, possibly due to unstable training gradients resulting from the uncertainty of the subclass labels. 
\par To make use of superclass labels without affecting the training of the subclass classification head, we propose a simple yet effective method called \textbf{Prior Concatenation (PC)}: as shown in Fig. \ref{fig:Network} (a), we concatenate predicted superclass logit scores $S_c(x^l)$ onto the feature maps $F(x^l)$ and then perform subclass segmentation. The intuition behind this operation is that by concatenating the predicted superclass probabilities with feature maps, the network is able to leverage the prior knowledge of the superclass distribution and focus more on learning the fine-grained features for better discrimination among subclasses.

\subsubsection{Separate Normalization}
\par Intuitively, 
given sufficient superclass labels in supervised learning, the superclassification head tends to reduce feature distance among samples within the same superclass, which conflicts with the goal of increasing the distance between subclasses within the same superclass. To alleviate this issue, we aim to enhance the internal diversity of the distribution within the same superclass while preserving the discriminative features among superclasses.

\par To achieve this, we propose \textbf{Separate Normalization(SN)} to separately process feature maps belonging to hierarchical foreground and background divided by superclass labels. As a superclass and the subclasses within share the same background, the original conflict between classifiers is transferred to finding the optimal transformations that separate foreground from background, enabling the network to extract class-specific features while keeping the features inside different superclasses well-separated. 

Our framework is shown in Fig. \ref{fig:Network} (b). First, we use Batch Norm layers\cite{ioffe2015batch} to perform separate affine transformations on the original feature map. The transformed feature maps, each representing a semantic foreground and background, are then passed through a convolution block for feature extraction before further classification. The classification process is coherent with the semantic meaning of each branch. Namely, the foreground branch includes a superclassifier and a subclassifier that classifies the superclass and subclass foreground, while the background branch is dedicated solely to classify background pixels. Finally, two separate network branches are jointly supervised by segmentation loss on super- and subclass labels. The aforementioned prior concatenation continues to take effect by concatenating predicted superclass logits on the inputs of subclassifier.

\subsubsection{HierarchicalMix}
Given the scarcity of subclass labels, we intend to maximally exploit the existent subclass supervision to guide the segmentation of coarsely labeled samples. Inspired by GuidedMix \cite{Peng2022}, which provides consistent knowledge transfer between similar labeled and unlabeled images with pseudo labeling, we propose \textbf{HierarchicalMix(HM)} to generate robust pseudo supervision. Nevertheless, GuidedMix relies on image distance to select similar images and performs a whole-image mixup, which loses focus on the semantic meaning of each region within an image. We address this limitation by exploiting the additional superclass information for a more targeted mixup. This information allows us to fuse only the semantic foreground regions, realizing a more precise transfer of foreground knowledge. A detailed pipeline of HierarchicalMix is described below.

\begin{figure}[tbp]
\centering
% {left, bottom, right, top}
\includegraphics[page=1, trim={3cm, 5.3cm, 5.5cm, 2cm}, clip, scale=0.45]{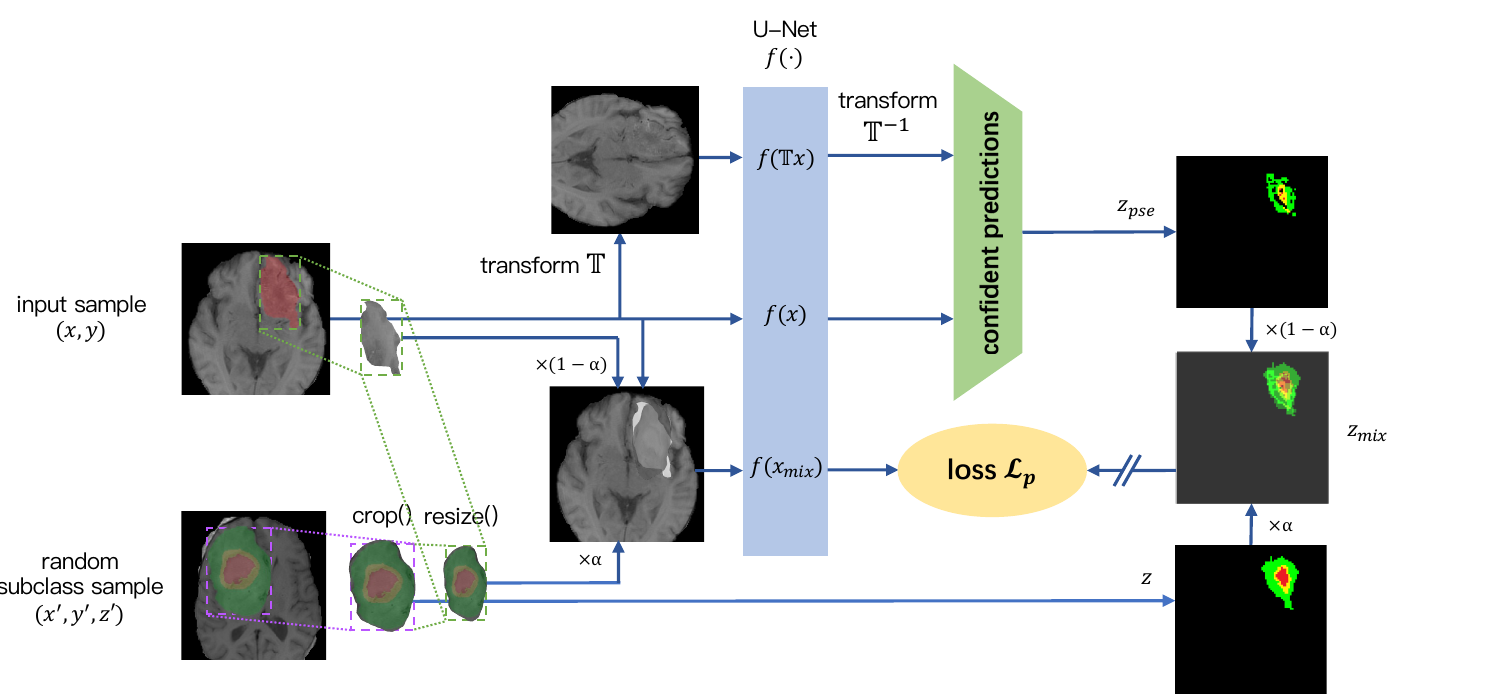}
\caption{The framework of $HierarchicalMix$. This process is adopted at training time to pair each coarsely labeled image $x$ with its mixed image $x_{mix}$ and pseudo subclass label $z$. ``$\sslash$'' represents the cut of gradient backpropagation.}
\label{fig:mixup}
\end{figure}

As shown in Fig. \ref{fig:mixup}, for each sample $(x,y)$ in the dataset that does not have subclass labels, we pair it with a randomly chosen fine-labeled sample $(x',y',z')$. First, we perform an random rotation and flipping $\mathbb{T}$ on $(x,y)$ and feed both the original sample and the transformed sample $\mathbb{T}x$ into the segmentation network $f$. An indirect segmentation of $x$ is obtained by performing the inverse transformation $\mathbb{T}^{-1}$ on the segmentation result of $\mathbb{T}x$. A transform-invariant pseudo subclass label map $z_{pse}$ is generated according to the following scheme: Pixel $(i,j)$ in $z_{pse}$ is assigned a valid subclass label index $(z_{pse})_{i,j}=f(x)_{i,j}$ only when $f(x)_{i,j}$ agrees with $[\mathbb{T}^{-1}f(\mathbb{T}x)]_{i,j}$ with a high confidence $\tau$ as well as $f(x)_{i,j}$ and $x_{i,j}$ both belong to the same superclass label.

Next, we adopt image mixup by cropping the bounding box of foreground pixels in $x'$, resizing it to match the size of foreground in $x$, and linearly overlaying them by a factor of $\alpha$ on $x$. This semantically mixed image $x_{mix}$ has subclass labels $z=\textnormal{resize}(\alpha\cdot z')$ from the fine-labeled image $x'$. Then, we pass it through the network to obtain a segmentation result $f(x_{mix})$. This segmentation result is supervised by the superposition of the pseudo label map $z_{pse}$ and subclass labels $z$, with weighting factor $\alpha$: $\mathcal{L}_p=\mathcal{L}(f(x_{mix}), \alpha\cdot z+(1-\alpha)\cdot z_{pse})$.

The intuition behind this framework is to simultaneously leverage the information from both unlabeled and labeled data by incorporating a more robust supervision from transform-invariant pseudo labels. While mixing up only the semantic foreground provides a way of exchanging knowledge between similar foreground objects while lifting the confirmation bias in pseudo labeling \cite{Eric2019}.

\section{Experiments}
\subsubsection{Dataset and preprocessing}
We conduct all experiments on two public datasets. The first one is the \textbf{ACDC}\footnote{\href{https://www.creatis.insa-lyon.fr/Challenge/acdc/databases.html}{https://www.creatis.insa-lyon.fr/Challenge/acdc/databases.html}} dataset~\cite{bernard2018deep}, which contains 200 MRI images with segmentation labels for left ventricle cavity (LV), right ventricle cavity (RV), and myocardium (MYO). Due to the large inter-slice spacing, we use 2D segmentation as in \cite{bai2017semi}. We adopt the processed data and the same data division in \cite{ssl4mis2020}, which uses 140 scans for training, 20 scans for validation and 40 scans for evaluation. During inference, predictions are made on each individual slice and then assembled into a 3D volume. The second is the \textbf{BraTS2021}\footnote{\href{http://braintumorsegmentation.org/}{http://braintumorsegmentation.org/}} dataset~\cite{baid2021rsna}, which consists of 1251 mpMRI scans with an isotropic 1 mm$^3$ resolution. Each scan includes four modalities (FLAIR, T1, T1ce, and T2), and is annotated for necrotic tumor core (TC), peritumoral edematous/invaded tissue (PE), and the GD-enhancing tumor (ET). We randomly split the dataset into 876, 125, and 250 cases for training, validation, and testing, respectively. For both datasets, image intensities are normalized to values in [0, 1] and the foreground superclass is defined as the union of all foreground subclasses for both datasets.

\subsubsection{Implementation details and evaluation metrics}
To augment the data during training, we randomly cropped the images with a patch size of $256 \times 256$ for the ACDC dataset and $96 \times 96 \times 96$ for the BraTS2021 dataset. The model loss $\mathcal{L}$ is set by adding the losses from Cross Entropy Loss and Dice Loss. % The weighing factor $\alpha$ in HierarchicalMix section is chosen to be 0.5, while $\tau$ linearly decreases from 1 to 0.4 during the training process.

We trained the model for 40,000 iterations using SGD optimizer with a 0.9 momentum and a linearly decreasing learning rate that starts at 0.01 and ends with 0. We used a batch size of 24 for the ACDC dataset and 4 for the BraTS2021 dataset, where half of the samples are labeled with subclasses and the other half only labeled with superclasses. More details can be found in the supplementary materials. To evaluate the segmentation performance, we used two widely-used metrics: the Dice coefficient ($DSC$) and 95\% Hausdorff Distance ($HD_{95}$). The confidence factor $\tau$ mentioned in HierarchicalMix starts at 1 and linearly decays to 0.4 throughout the training process, along with a weighting factor $\alpha$ sampled according to the uniform distribution on $[0.5, 1]$.

\subsubsection{Performance comparison with other methods}
To evaluate the effectiveness of our proposed method, we firstly trained two \textbf{U-Net} models \cite{ronneberger2015u} to serve as upper and lower bounds of performance. The first U-Net was trained on the complete subclass dataset $\{(x^l,y^l,z^l)\}_{l=1}^N$, while the second was trained on its subset $\{(x^l,y^l,z^l)\}_{l=1}^n$. Then, we compared our method with the following four methods, all of which were trained using $n$ subclass labels and $N$ superclass labels: \textbf{Modified U-Net (Mod)}: This method adds an additional superclass classifier alongside the subclass classifier in the U-Net.
\textbf{Negative Learning (NL)}: This method incorporates superclass information into the loss module by introducing a separate negative learning loss in the original U-Net. This additional loss penalizes pixels that are not segmented as the correct superclass.
\textbf{Cross Pseudo Supervision (CPS)}~\cite{chen2021semi}: This method simulates pseudo supervision by utilizing the segmentation results from two models with different parameter initializations, and adapts their original network to the Modified U-Net architecture.
\textbf{Uncertainty Aware Mean Teacher (UAMT)}\cite{yu2019uncertainty}: This method modifies the classical mean teacher architecture\cite{tarvainen2017mean} by adapting the teacher model to learn from only reliable targets while ignoring the rest, and also adapts the original network to the Modified U-Net architecture.

\begin{table}[htbp]
\begin{center}
\caption{Mean Dice Score (\%, left) and $HD_{95}$ (mm, right) of different methods on ACDC and BraTS2021 datasets. Sup. and Sub. separately represents the number of data with superclass and subclass annotations in the experiments. `$\_$' means the result of our proposal is significantly better than the closet competitive result (p-value < 0.05). The standard deviations of each metric are recorded in the supplementary materials.}
\label{table:results}
\scalebox{0.68}{
\begin{tabular}{c|cc|cccc|cc|cccc} 
\toprule
\multirow{2}{*}{\textbf{Method}} & \multicolumn{6}{c|}{\textbf{ACDC}} & \multicolumn{6}{c}{\textbf{BraTS2021}}\\
\cline{2-13}
&Sup.&Sub.&RV&MYO&LV&Avg.&Sup.&Sub.&TC&PE&ET&Avg.\\
\midrule
U-Net&0&3&36.6, 61.5&51.6, 20.7&57.9, 26.2&48.7, 36.2&0&10&57.5, 16.6&68.8, 22.9&74.7, 12.4&67.0, 17.3\\
U-Net&0&140&90.6, 1.88&89.0, 3.59&94.6, 3.60&91.4, 3.02&0&876&75.8, 4.86&82.2, 5.87&83.6, 2.48&80.6, 4.40\\
\hline
Mod&140&3&83.1, 11.1&80.7, 6.12&83.1, 14.7&82.3, 10.6&876&10&60.3, 7.69&76.2, 7.70&80.2, 4.97&72.3, 6.79\\
NL\cite{kim2019nlnl}&140&3&61.0, 18.8&68.6, 13.7&81.5, 19.5&70.4, 17.3&876&10&59.5, 10.5&75.2, 8.35&76.8, 6.34&70.5, 8.40\\
CPS\cite{chen2021semi}&140&3&80.2, 9.54&80.3, 3.17&86.3, \textbf{4.21}&82.3, 5.64&876&10&62.9, 7.02&78.3, 7.08&80.8, 4.91&74.0, 6.24\\
UAMT\cite{yu2019uncertainty}&140&3&79.4, 7.81&77.7, 5.87&85.5, 8.16&80.9, 7.28&876&10&60.8, 9.84&78.4, 7.11&80.1, 4.24&73.3, 7.06\\
Ours&140&3&\underline{\textbf{87.2}}, \underline{\textbf{1.84}}&\underline{\textbf{84.6}}, \textbf{2.70}&\underline{\textbf{90.1}}, 4.44&\underline{\textbf{87.3}}, \underline{\textbf{2.99}}&876&10&\textbf{65.5}, \textbf{6.90}&\textbf{79.9}, \textbf{6.38}&\textbf{80.8}, \textbf{3.59}&\textbf{75.4}, \textbf{5.62}\\
\bottomrule
\end{tabular}}
\end{center}
\end{table}
\vspace{-0.3cm}
\par The quantitative results presented in Table \ref{table:results} reveal that all methods that utilize additional superclass annotations outperformed the baseline method, which involved training a U-Net using only limited subclass labels. However, the methods that were specifically designed to utilize superclass information or explore the intrinsic structure of the subclass data, such as NL, CPS, and UAMT, did not consistently outperform the simple Modified U-Net. In fact, these methods sometimes performed worse than the simple Modified U-Net, indicating the difficulty of utilizing superclass information effectively. In contrast, our proposed method achieved the best performance among all compared methods on both the ACDC and BraTS2021 datasets. Specifically, our method attained an average Dice score of 87.3\% for ACDC and 75.4\% for BraTS2021, outperforming the closest competitor by 5.0\% and 1.4\%, respectively.

\begin{table}[tbp]
\begin{center}
\caption{Mean Dice Score (\%, left) and $HD_{95}$ (mm, right) of ablation studies on ACDC and BraTS2021 datasets ($mixup$ and $pseudo$ in HM column separately stands for using solely image mixup and pseudo-labeling to achieve better data utilization).}
\label{table:ablations}
\scalebox{0.68}{
\begin{tabular}{ccc|cc|cccc|cc|cccc} 
\toprule
\multirow{2}{*}{HM} & \multirow{2}{*}{PC} & \multirow{2}{*}{SN}& \multicolumn{6}{c|}{\textbf{ACDC}} & \multicolumn{6}{c}{\textbf{BraTS2021}}\\
\cline{4-15}
&&&Sup.&Sub.&RV&MYO&LV&Avg.&Sup.&Sub.&TC&PE&ET&Avg.\\
\midrule
&&&140&3&83.1, 11.1&80.7, 6.12&83.1, 14.7&82.3, 10.6&876&10&60.3, 7.69&76.2, 7.70&80.2, 4.97&72.3, 6.79\\
$\checkmark$&&&140&3&85.9, 2.55&83.6, 3.70&89.8, 5.15&86.5, 3.80&876&10&65.0, 8.00&77.0, 7.47&80.6, 3.74&74.2, 6.40\\
&$\checkmark$&&140&3&80.0, 8.06&80.4, 6.63&87.9, 5.07&82.8, 6.58&876&10&61.6, 7.00&77.3, 6.89&80.4, 6.01&73.1, 6.63\\
&&$\checkmark$&140&3&79.0, 3.32&81.2, 3.69&88.6, 4.43&82.9, 3.82&876&10&63.5, 9.03&78.9, 6.29&80.2, 4.45&74.2, 6.59\\
$\checkmark$&$\checkmark$&&140&3&85.1, 1.86&81.4, 4.29&87.3, 5.55&84.6, 3.90&876&10&65.1, 7.93&78.4, 6.86&78.3, 3.97&73.9, 6.25\\
$\checkmark$&&$\checkmark$&140&3&\textbf{87.6}, 2.81&83.8, \textbf{2.06}&89.9, 2.87&87.1, \textbf{2.58}&876&10&65.7, 7.56&79.6, 6.68&\textbf{81.4}, 4.25&75.5, 6.16\\
&$\checkmark$&$\checkmark$&140&3&84.7, 5.26&84.1, 2.53&89.3, \textbf{2.79}&86.0, 3.53&876&10&64.4, 7.96&79.5, 6.41&79.5, 5.07&74.4, 6.48\\
$mixup$&$\checkmark$&$\checkmark$&140&3&82.9, 5.42&80.6, 4.18&86.8, 6.06&83.5, 5.22&876&10&\textbf{66.2}, 6.90&79.6, 6.26&80.9, 4.19&\textbf{75.6}, 5.79\\
$pseudo$&$\checkmark$&$\checkmark$&140&3&78.8, 12.2&80.1, 7.66&84.3, 7.71&81.1, 9.20&876&10&62.4, 11.1&77.9, 6.55&80.0, 7.09&73.5, 8.24\\
$\checkmark$&$\checkmark$&$\checkmark$&140&3&87.2, \textbf{1.84}&\textbf{84.6}, 2.70&\textbf{90.1}, 4.44&\textbf{87.3}, 2.99&876&10&65.5, \textbf{6.90}&\textbf{79.9}, \textbf{6.38}&80.8, \textbf{3.59}&75.4, \textbf{5.62}\\
\hline
$\checkmark$&$\checkmark$&$\checkmark$&140&6&86.6, 1.20&84.7, 1.87&90.9, 4.23&87.4, 2.44&876&20&70.7, 7.45&81.2, 6.08&82.2, 3.58&78.0, 5.70\\
$\checkmark$&$\checkmark$&$\checkmark$&140&9&86.1, 1.78&85.7, 1.92&90.8, 4.15&87.6, 2.62&876&30&71.4, 6.15&81.4, 5.84&82.5, 3.25&78.5, 5.08\\
\hline
\multicolumn{3}{c|}{UNet}&0&140&90.6, 1.88&89.0, 3.59&94.6, 3.60&91.4, 3.02&0&876&75.8, 4.86&82.2, 5.87&83.6, 2.48&80.6, 4.40\\
\bottomrule
\end{tabular}}
\end{center}
\end{table}

\subsubsection{Ablation studies} 
In this study, we performed comprehensive ablation studies to analyze the contributions of each component and the performance of our method under different numbers of images with subclass annotations. The performance of each component is individually evaluated, and is listed in Table~\ref{table:ablations}.

\par Each component has demonstrated its effectiveness in comparison to the naive modified U-Net method. Moreover, models that incorporate more components generally outperform those with fewer components. The effectiveness of the proposed HierarchicalMix is evident from the comparisons made with models that use only image mixup or pseudo-labeling for data augmentation, while the addition of Separate Normalization consistently improves the model performance. Furthermore, our method was competitive with a fully supervised baseline, achieving comparable results with only 6.5\% and 3.4\% subclass annotations on ACDC and BraTS2021.

\section{Conclusion}
In this work, we proposed an innovative approach to address the problem of efficient subclass segmentation in medical images, where limited subclass annotations and sufficient superclass annotations are available. To the best of our knowledge, this is the first work specifically focusing on this problem. Our approach leverages the hierarchical structure of categories to design network architectures and data generation methods that enable the network to distinguish between subclass categories more easily. Specifically, we introduced a Prior Concatenation module that enhances confidence in subclass segmentation by concatenating predicted logits from the superclass classifier, a Separate Normalization module that stretches the intra-class distance within the same superclass to facilitate subclass segmentation, and a HierarchicalMix model that generates high-quality pseudo labels for unlabeled samples by fusing only similar superclass regions from labeled and unlabeled images. Our experiments on the ACDC and BraTS2021 datasets demonstrated that our proposed approach outperformed other compared methods in improving the segmentation accuracy. Overall, our proposed method provides a promising solution for efficient fine-grained subclass segmentation in medical images.

\bibliographystyle{splncs04}
\bibliography{ref}

\newpage
\section{Supplementary Materials}

\begin{table}[htb]
\begin{center}
\caption{Training parameters on ACDC and BraTS2021 datasets.}
\label{table:param1}
\begin{tabular}{c|cc}
\toprule
Parameter&ACDC&BraTS2021\\
\midrule
GPU&\multicolumn{2}{c}{One Nvidia 3090 GPU}\\
Program language&\multicolumn{2}{c}{Python 3.7.13}\\
Training framework&\multicolumn{2}{c}{PyTorch 1.10.0}\\
Base learning rate&\multicolumn{2}{c}{0.01}\\
Total iterations&\multicolumn{2}{c}{40000}\\
Batch size& 24 & 4\\
Patch size&$256\times256$&$96\times96\times96$\\
Memory Usage & 8GB &10GB\\
\bottomrule
\end{tabular}
\end{center}
\end{table}

\begin{table}[h]
\begin{center}
\caption{Training time (days) of different methods on ACDC and BraTS2021 datasets.}
\label{table:param2}
\begin{tabular}{c|cccccc}
\toprule
Dataset&U-Net&Mod&NL&CPS&UAMT&Ours\\
\midrule
ACDC&0.12&0.22&0.14&0.31&0.21&0.23\\
BraTS2021 &0.78&1.77&0.80&3.83&2.60&2.56 \\
\bottomrule
\end{tabular}
\end{center}
\end{table}

\begin{table}[H]
\begin{center}
\caption{Mean Dice Score (\%, left) and $HD_{95}$ (mm, right) and their standard deviations of different methods on ACDC dataset. Sup. and Sub. separately represents the number of data with superclass and subclass annotations in the experiments. `$\_$' means the result of our proposal is significantly better than the closet competitive result (p-value < 0.05).}
\label{table:results1}
\scalebox{0.75}{
\begin{tabular}{c|cc|cccc} 
\toprule
\textbf{Method}&Sup.&Sub.&LV&MYO&RV&Avg.\\
\midrule
U-Net&0&3&36.6$\pm$29.1, 61.5$\pm$30.0&51.6$\pm$30.0, 20.7$\pm$23.6&57.9$\pm$34.3, 26.2$\pm$32.0&48.7$\pm$30.0, 36.2$\pm$2.2\\
U-Net&0&140&90.6$\pm$6.32, 1.88$\pm$4.12&89.0$\pm$2.96, 3.59$\pm$10.6&94.6$\pm$3.59, 3.60$\pm$10.1&91.4$\pm$2.92, 3.02$\pm$5.77\\
\hline
%Ours w./o. data augmentation&0.8292&0.7407\\
%Ours w./o. network optimization&0.8646&0.7419\\
Mod&140&3&83.1$\pm$9.33, 11.1$\pm$24.6&80.7$\pm$6.43, 6.12$\pm$6.98&83.1$\pm$13.3, 14.7$\pm$13.7&82.3$\pm$8.00, 10.6$\pm$10.2\\
NL&140&3&61.0$\pm$26.0, 18.8$\pm$16.3&68.6$\pm$20.6, 13.7$\pm$14.4&81.5$\pm$14.1, 19.5$\pm$19.8&70.4$\pm$18.3, 17.3$\pm$12.7\\
CPS&140&3&80.2$\pm$9.99, 9.54$\pm$12.9&80.3$\pm$7.40, 3.17$\pm$4.47&86.3$\pm$10.4, \textbf{4.21}$\pm$10.3&82.3$\pm$7.28, 5.64$\pm$5.68\\
UAMT&140&3&79.4$\pm$11.7, 7.81$\pm$13.7&77.7$\pm$9.73, 5.87$\pm$8.74&85.5$\pm$11.2, 8.16$\pm$10.5&80.9$\pm$9.21, 7.28$\pm$8.56\\
Ours&140&3&\underline{\textbf{87.2}$\pm$7.44}, \underline{\textbf{1.84}$\pm$1.84}&\underline{\textbf{84.6}$\pm$4.97}, \textbf{2.70}$\pm$4.87&\underline{\textbf{90.1}$\pm$9.25}, 4.44$\pm$10.4&\underline{\textbf{87.3}$\pm$5.33}, \underline{\textbf{2.99}$\pm$4.40}\\
\bottomrule
\end{tabular}}
\end{center}
\end{table}

\begin{table}[tbp]
\begin{center}
\caption{Mean Dice Score (\%, left) and $HD_{95}$ (mm, right) and their standard deviations of different methods on BraTS2021 dataset. Sup. and Sub. separately represents the number of data with superclass and subclass annotations in the experiments. `$\_$' means the result of our proposal is significantly better than the closet competitive result (p-value < 0.05).}
\label{table:results2}
\scalebox{0.75}{
\begin{tabular}{c|cc|cccc} 
\toprule
\textbf{Method}&Sup.&Sub.&TC&PE&ET&Avg.\\
\midrule
U-Net&0&10&57.5$\pm$29.1, 16.6$\pm$22.5&68.8$\pm$22.1, 22.9$\pm$26.2&74.7$\pm$21.6, 12.4$\pm$21.8&67.0$\pm$20.0, 17.3$\pm$19.2\\
U-Net&0&876&75.8$\pm$23.9, 4.86$\pm$9.26&82.2$\pm$15.8, 5.87$\pm$9.72&83.6$\pm$17.4, 2.48$\pm$4.16&80.6$\pm$14.9, 4.40$\pm$5.84\\
\hline
Mod&876&10&60.3$\pm$28.5, 7.69$\pm$7.73&76.2$\pm$18.3, 7.70$\pm$9.01&80.2$\pm$17.6, 4.97$\pm$8.47&72.3$\pm$17.7, 6.79$\pm$6.11\\
NL&876&10&59.5$\pm$28.4, 10.5$\pm$15.1&75.2$\pm$15.8, 8.35$\pm$11.4&76.8$\pm$20.5, 6.34$\pm$12.6&70.5$\pm$16.2, 8.40$\pm$9.98\\
CPS&876&10&62.9$\pm$26.8, 7.02$\pm$7.97&78.3$\pm$18.4, 7.08$\pm$10.1&80.8$\pm$17.8, 4.91$\pm$9.79&74.0$\pm$17.0, 6.24$\pm$7.90\\
UAMT&876&10&60.8$\pm$26.7, 9.84$\pm$10.7&78.4$\pm$17.4, 7.11$\pm$9.97&80.1$\pm$18.0, 4.24$\pm$6.40&73.3$\pm$16.3, 7.06$\pm$6.18\\
Ours&876&10&\textbf{65.5}$\pm$26.6, \textbf{6.90}$\pm$9.63&\textbf{79.9}$\pm$17.3, \textbf{6.38}$\pm$9.32&\textbf{80.8}$\pm$18.6, \textbf{3.59}$\pm$8.34&\textbf{75.4}$\pm$16.5, \textbf{5.62}$\pm$6.45\\
\bottomrule
\end{tabular}}
\end{center}
\end{table}

\end{document}